# Phase 1 Implementation of LLM-generated Discharge Summaries showing high Adoption in a Dutch Academic Hospital

Nettuno Nadalini, BSc[1]
Tarannom Mehri, BSc[1]
Anne H. Hoekman, MD, LLM[1]
Katerina Kagialari, BSc[1]
Job N. Doornberg, MD, PhD[1,2,3,4,5,6]
Tom P. van der Laan, MD, PhD[1,7]
Jacobien H.F. Oosterhoff, MD, PhD[2]
Rosanne C. Schoonbeek, MD, PhD[1,7]
Charlotte M.H.H.T. Bootsma-Robroeks, MD, PhD[1,8]
*On behalf of the Generative AI Network (GAIN), part of the Applied AI Accelerator (A3) Lab*

**Affiliations**
1. Department of Health Information Office, Data & Digitalization, University Medical Center, Groningen, the Netherlands
2. Department of Trauma Surgery / Orthopedics, University Medical Center, Groningen, the Netherlands
3. Department of Orthopaedic Surgery, Massachusetts General Hospital, Boston, MA, USA
4. Harvard Medical School Orthopaedic Trauma Initiative, Boston, MA, USA
5. Department of Orthopaedic Surgery and Sports Medicine, Amsterdam UMC location University of Amsterdam, Meibergdreef 9, Amsterdam, The Netherlands
6. Department of Orthopaedic Trauma, Flinders University Medical Centre, Adelaide, Australia
7. Department of Otolaryngology – Head and Neck Surgery, University Medical Center, Groningen, the Netherlands
8. Department of Pediatrics, Pediatrics Nephrology, Beatrix Children's Hospital, University Medical Center, Groningen, the Netherlands

**Corresponding Author**
Nettuno Nadalini, BSc
Email address: n.nadalini@umcg.nl
Telephone number: +352 621 529 141
ORCID: 0009-0009-6296-3156
Department of Health Information Office, Information Management Healthcare
University Medical Center Groningen, University of Groningen, the Netherlands.
Hanzeplein 1, 9713 GZ Groningen, the Netherlands

## Abstract

Writing discharge summaries to transfer medical information is an important but time-consuming process that can be assisted by Large Language Models (LLMs). This prospective mixed methods pilot study evaluated an Electronic Health Record (EHR)-integrated LLM to generate discharge summaries drafts. In total, 379 discharge summaries were generated in clinical practice by 21 residents and 4 physician assistants during 9 weeks in our academic hospital.

LLM-generated text was copied in 58.5% of admissions, and identifiable LLM content could be traced to 29.1% of final discharge letters. Notably, 86.9% of users self-reported a reduction in documentation time, and 60.9% a reduction in administrative workload. Intent to use after the pilot phase was high (91.3%), supporting further implementation of this use-case. Accurately measuring the documentation time of users on discharge summaries remains challenging, but will be necessary for future extrinsic evaluation of LLM-assisted documentation.

## Introduction

Discharge summaries, present in discharge letters, are an important communication tool transferring medical information between healthcare professionals (HCPs) [1,2]. The population of multi-morbid patients necessitating care across multiple institutions is increasing, highlighting the need for timely and adequate discharge summaries [3,4]. To this end, every discharge needs to be accompanied by a (provisional) digital discharge letter within 24 to 48 hours according to national guidelines (https://richtlijnendatabase.nl/). However, ensuring high quality, correctness, completeness and timeliness in discharge summaries is a time-consuming process, necessitating up to 10% of daily working time [5,6]. By consequence, discharge summaries are often missing or sent too late and contribute to medical error and disputes between HCPs [2,7]. Since their introduction in 2006 (in the Netherlands), Electronic Health Record systems (EHRs) have not been used to their full potential [8,9]. While ensuring better patient safety, this has come at the cost of increased administrative workload for HCPs, often performed outside working hours [6,10–14]. Multiple studies report that HCPs spend more time in the EHR than with their patients, clinical documentation tasks being the most important contributor [10,12,13]. This increased administrative burden is associated with a lower quality of care, increased physician burnout rates and decreased job satisfaction [15–18].

Generative Artificial Intelligence (GenAI), by means of Large Language Models (LLMs), has the potential to reduce administrative burden for HCPs [19–21]. LLMs can process and interpret human language through a process known as 'deep learning'. With pre-training and fine tuning, these technologies are capable of accurately summarising medical information extracted from electronic medical records, sometimes outperforming HCPs [22–24]. While challenging, previous research has proven the potential of LLMs to quickly generate qualitatively adequate discharge summaries [25–32]. However, transforming potential into practice remains challenging, and studies reporting on clinical value and user acceptance remain scarce [33–35].

To address this, we performed a prospective mixed methods pilot study on the first implementation phase of an LLM-driven EHR-integrated discharge summary generator in daily Dutch clinical practice. Adoption and user experience of the LLM were the main outcomes, and potential time benefits were explored for extrinsic validation. To the best of our knowledge, this is the first multi-department study reporting on the deployment of LLM-generated discharge summaries in real-world clinical practice [29,31,32,34–37].

## Results

### Adoption

The LLM was used to generate 379 discharge summary drafts for 253 admissions (1.5 generations per admission, IQR=1) across 10 departments in 9 weeks. The LLM-generated draft was copied 148 times. 58.5% of all admissions with an LLM-generated draft had it copied. The overall generations-to-copy ratio was 2.7. LLM-generated text was present in at least 66 discharge letters, as shown by figure 1.

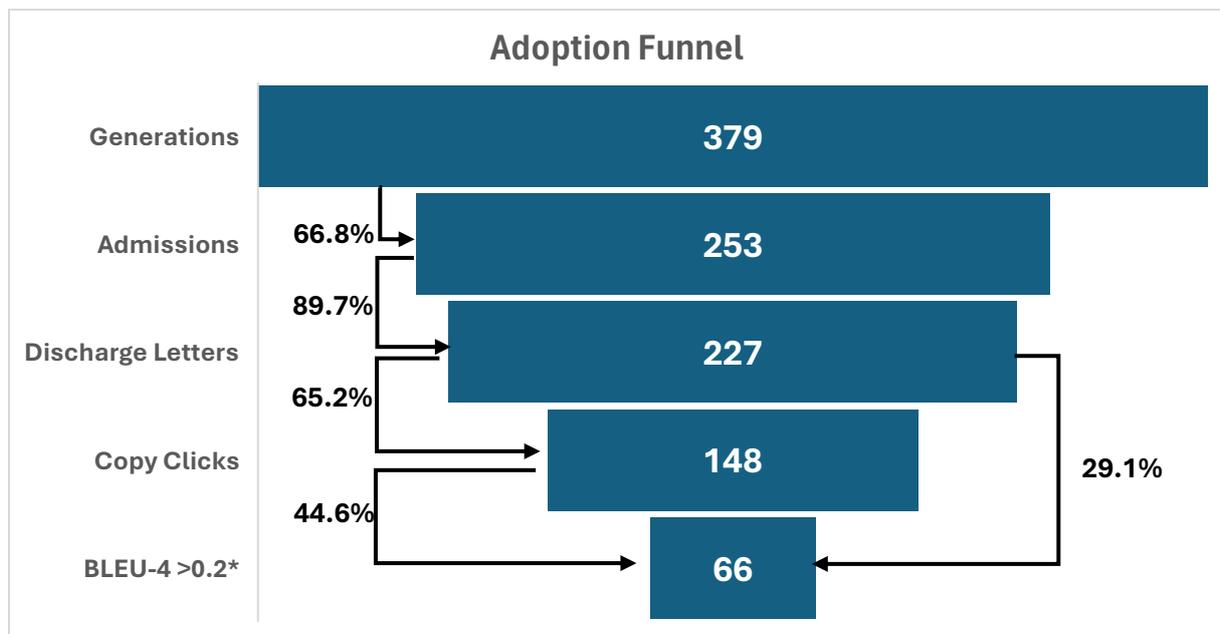

*Figure 1: Funnel Diagram quantifying the stages of adoption. Numbers in boxes are absolute, percentages next to arrows indicate retainment. *This reflects the least amount of discharge letters containing LLM-generated text as estimated by BLEU-scores.*

BLEU-scores were used to quantify the amount of LLM-generated text copied into the discharge summary of the final discharge letter. BLEU-scores were calculated for 227 admissions, as 26 discharge summaries could not be retrieved. The BLEU-2 score was the closest estimator of the amount of LLM-generated text copied into the final discharge summary but tended to overestimate the presence of LLM-generated text at low scores. The BLEU-4 score was more reliable in this regard, guaranteeing the presence of LLM-generated text above BLEU-4 scores of 0.2. This is exemplified by figures 2 and 3, and supplementary figures 1-6. Figure 3 shows that LLM-generated text can also be found in some letters with a BLEU-4 score below 0.2.

To achieve a BLEU-4 score above 0.2, users needed an average of 1.7 (IQR=1) generations. High BLEU scores indicated the LLM-generated text was good enough to be used with little editing, as seen in figure 2. Cutoffs of 0.5 and 0.8 were arbitrarily added to visualize this in figure 4. 66 (29.1%) discharge summaries had a BLEU-4 score above 0.2 and 31 (13.7%) had a BLEU-4 score above 0.5. Figure 4 shows a similar proportion of BLEU-2 scores above 0.5 (37, 16.3%), but a higher proportion of BLEU-2 scores above 0.2 (107, 47.1%) when compared to BLEU-4 scores.

| [Name], [age], known with granulomatosis with polyangiitis (GPA) with airway involvement and diabetes mellitus type 2, was admitted because of clinical and pulmonary functional deterioration despite ongoing treatment for infectious complications and tapering of immunosuppressants. During the admission, on basis of clinical picture, pulmonary function, BAL (Pseudomonas, Proteus, Aspergillus), and airway biopsy (diffuse inflammation, eosinophilia), an exacerbation of GPA and an active infectious component were established, in multidisciplinary consultation with rheumatology, pulmonology and nephrology. Rituximab was discontinued and azathioprine started, with optimization of the infectious treatment including azithromycin and tobramycin inhalations; IVIG was added because of hypogammaglobulinemia. Voriconazole was continued and the level will be determined. Her oxygen requirement was evaluated and home supply arranged, given subjective improvement of dyspnoea with 2L O2 during exertion. Physiotherapy was applied for mobilization and sputum clearance, with increase of mobility through walker. Dietetics was consulted because of [weight] kg weight loss in 4 months, with advice focused on energy- and protein-enriched nutrition; her weight at discharge was [weight] kg. During the admission her diabetes remained stable and no acute infections or other complications were reported. Her resuscitation status was discussed and recorded as do-not-resuscitate. | Above-mentioned [age]-old patient, known with granulomatosis with polyangiitis (GPA) with airway involvement and diabetes mellitus type 2, was admitted because of functional and pulmonary functional deterioration despite ongoing treatment for infectious component and tapering of immunosuppressants. During the admission patient was discussed in a multidisciplinary consultation with rheumatology, pulmonology and nephrology. Rituximab was discontinued and azathioprine started, with optimization of the infectious treatment including start azithromycin and continuation tobramycin inhalations; IVIG was added because of hypogammaglobulinemia of which she received the first dose during admission on [date]. Voriconazole was continued and the level will be determined. Her oxygen requirement was evaluated and home supply arranged, given subjective improvement of dyspnoea with 2L O2 during exertion. Physiotherapy was applied for mobilization and sputum clearance, with increase of mobility through walker. Dietetics was consulted because of [weight] kg weight loss in 4 months, with advice focused on energy- and protein-enriched nutrition; her weight at discharge was [weight] kg. During the admission her diabetes remained stable and no acute infections or other complications were reported. |
|---|---|

***Figure 2: Pair of LLM-generated discharge summary (left) and the discharge summary sent in the final letter (right).*** *The texts were de-identified and translated from Dutch to English word-by-word to maintain semantic structure and are not grammatically correct. The highlighted text is identical between the two. The manual score was obtained by dividing the number of identical words by the total number of words in the discharge summary included in the final discharge letter. BLEU-2 = 0.73; BLEU-4 = 0.70, manual score = 0.88.*

| *[Name], [age], with a history of hypothyroidism, ==PKD, bilateral nephrectomy, kidney transplantation in [date]. and recurrent urinary tract infections, was admitted== on [date] because of a urinary tract infection. The diagnosis urinary tract infection was confirmed on basis of clinical presentation and a positive urine culture for Klebsiella pneumoniae. ==Additionally there was suspicion of a pneumonia,== whereby on ==POCUS infiltrative abnormalities== were seen right basally, despite a clear X-thorax. Laboratory investigation showed an elevated CRP (maximum [value] mg/L), a decreasing creatinine (from [value] to [value] umol/L), and a stable Hb. Initially was started with meropenem, later changed to oral augmentin guided by clinical recovery. The prednisolone dosage was reduced to the home situation. Cellcept was stopped. The indwelling urinary catheter was removed. No additional oxygen was given after clinical recovery of the respiratory complaints. Fosfomycin maintenance was evaluated with a view to possible adjustment in the outpatient clinic.* <br> *During the admission there was no indication for cyst infection given the status after bilateral nephrectomy and the absence of liver cysts. The patient developed no diarrhoea and remained independent for daily tasks. There was a good clinical improvement, with decrease of the inflammatory parameters and recovery of the kidney function.* | *Above-mentioned [age]-old patient known with ==PKD== for which ==bilateral nephrectomy==, after which ==kidney transplantation in [date] and recurrent urinary tract infections was admitted== in connection with a urinary tract infection. ==Additionally there was a suspicion of a pneumonia==, given patient had dyspnoea and a ==POCUS== possibly did show ==infiltrative abnormalities==. This, however, was not confirmed on X-thorax. Was initially started with meropenem IV on basis of known resistance. The urine culture was positive for Klebsiella pneumoniae. On basis of the antibiogram was switched to augmentin with clinical recovery. There were no indications for an infected cyst (given the prior PKD). Patient previously had fosfomycin as maintenance treatment for the recurrent urinary tract infections. In the outpatient clinic in consultation with her own nephrologist did will be decided whether maintenance antibiotics will be restarted.* |
|---|---|

***Figure 3: Pair of LLM-generated discharge summary (left) and the discharge summary sent in the final letter (right).*** *The texts were de-identified and translated from Dutch to English word-by-word to maintain semantic structure and are not grammatically correct. The highlighted text is identical between the two. The manual score was obtained by dividing the number of identical words by the total number of words in the discharge summary included in the final discharge letter. BLEU-2 = 0.22; BLEU-4 = 0.08, manual score = 0.18.*

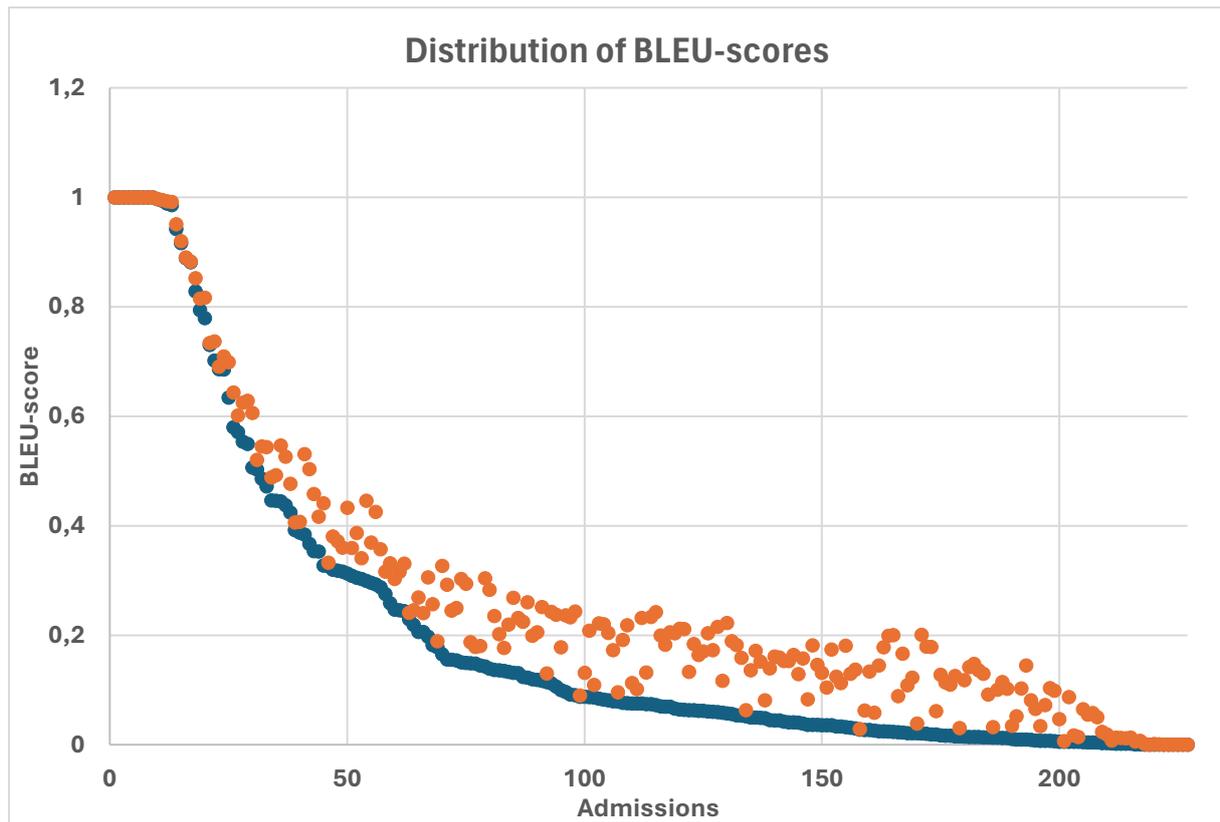

*Figure 4: Distribution of BLEU-scores. Scatter plot of the BLEU-4 (dark blue) and BLEU-2 (orange) scores plotted for 227 admissions. The admissions are sorted by BLEU-4 scores for interpretability, from high (left) to low (right).*

Adoption varied per department, as shown by table 1. The surgery department demonstrated a high generation count, low generations-to-copy ratio, and high BLEU-scores. The emergency department had the lowest adoption numbers.

| Department | Generations | Copy Clicks | Generations /Copy | BLEU-2 | BLEU-4 |
|---|---|---|---|---|---|
| Cardiology | 27 | 20 | 1.4 | 0.16 (0.24) | 0.06 (0.25) |
| Surgery | 50 | 25 | 2.0 | 0.52 (0.98) | 0.50 (0.99) |
| Internal Medicine | 70 | 26 | 2.7 | 0.22 (0.15) | 0.08 (0.16) |
| Paediatrics | 89 | 24 | 3.7 | 0.19 (0.26) | 0.05 (0.27) |
| Ear, Nose and Throat (ENT) Surgery | 21 | 5 | 4.2 | 0.18 (0.1) | 0.08 (0.07) |
| Pulmonology | 20 | 12 | 1.7 | 0.24 (0.22) | 0.10 (0.24) |
| Gastro-enterology | 31 | 24 | 1.3 | 0.20 (0.13) | 0.11 (0.11) |
| Neurology & Neurosurgery | 40 | 4 | 10.0 | 0.14 (0.1) | 0.03 (0.05) |
| Orthopaedics | 20 | 2 | 10.0 | 0.11 (0.14) | 0.09 (0.18) |
| Emergency Department | 11 | 6 | 1.8 | 0.08 (0.02) | 0.02 (0.01) |
| **Total** | **379** | **148** | **2.7** | **0.19 (0.22)** | **0.07 (0.26)** |

*Table 1: Adoption numbers per department. From left to right: number of successful generations achieved, copy clicks performed, generations-to-copy ratio, median BLEU-2 and BLEU-4 scores with the interquartile range in parentheses.*

**User Experience**

23 of 25 pilot users completed the survey after the study (92%). Time benefit was experienced by 86.9% of respondents and 60.9% experiencing some reduction in workload, all indicating perceived usefulness, Figure 5. Trust in the LLM's output varied, with 39.1% of respondents having trust to various degrees, 26.1% being neutral and 34.7% distrusting it to various degrees. Additionally, question 5 shows 70% of respondents would trust their colleagues with usage of the LLM.

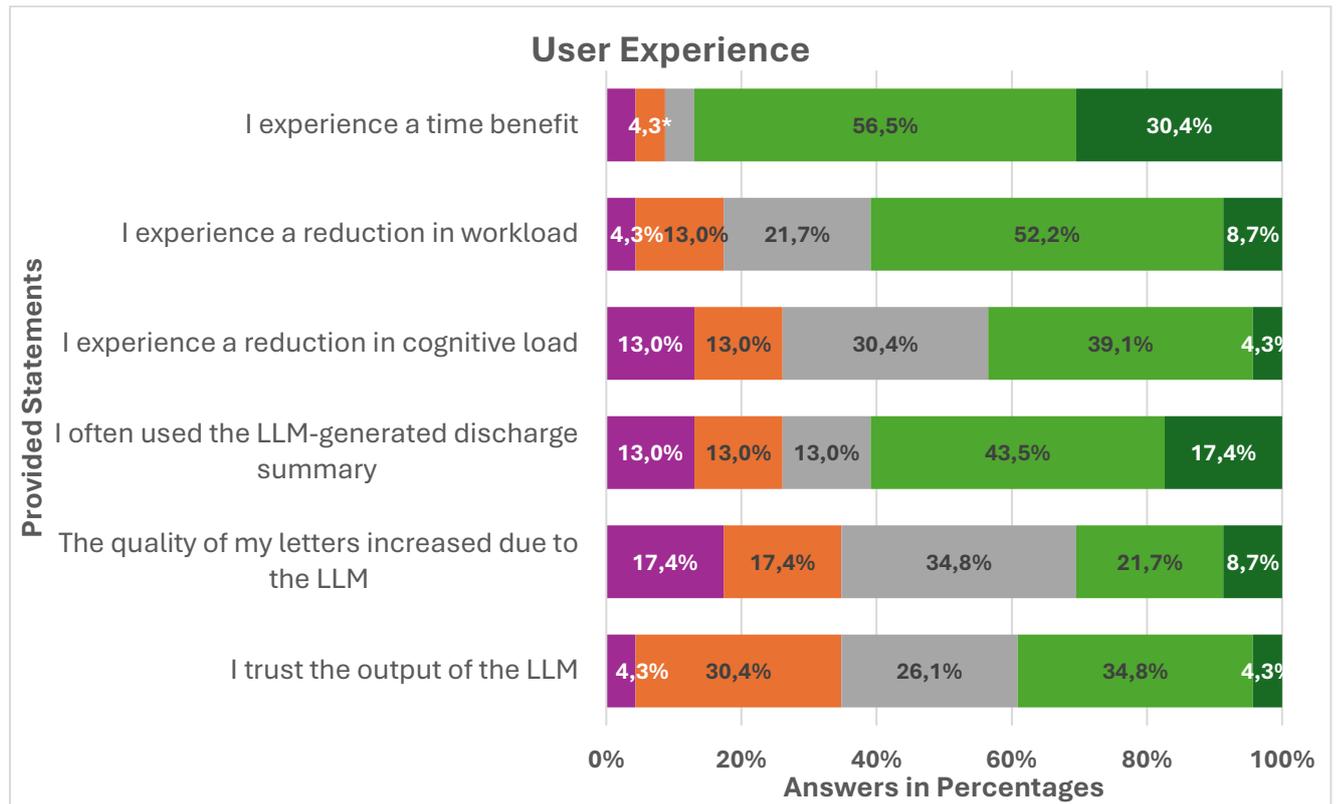

*Figure 5: Agreement with the given statements about the effects of the EHR-integrated LLM in percentages. In colours from left to right: disagree (purple), partly disagree (orange), neutral (grey), partly agree (light green), agree (dark green). *Identical values (4.3%) for disagree, partly disagree, and neutral are grouped for the first statement for the sake of clarity.*

When asked to rank features of the LLM, ease of use was ranked most valuable, followed by completeness, correctness and conciseness. According to 39.1% of respondents, using the EHR-integrated LLM was easy and immediately beneficial, while another 39.1% stated that they initially had to invest energy and time to experience a benefit. 21.8% Saw the potential of the LLM but did not personally experience any benefits, and none perceived neither potential nor benefits. Overall, 91.3% of respondents expressed intent to use beyond the study and 78.3% think the LLM is ready for further implementation.

**Potential Time Benefits**

All data was non-normally distributed. The documentation time of a singular user per admission (time-in-workflow) and the length of stay (LOS) of the inpatient admission were correlated (Pearson's ρ 0.429; $P$ <0.001).

Usage of the LLM significantly increased time-in-workflow when compared to groups without LLM usage, as exemplified by before and after ($P$ = 0.003) and before and LLM ($P$ = 0.000) comparisons. LOS was significantly higher in admissions with an LLM-generated draft ($P$ < 0.001 for before vs LLM; $P$ = 0.000 for no LLM vs LLM). The time between discharge and discharge letter sent (turnaround-time) significantly decreased after introduction of the LLM ($P$ = 0.043), but not in groups with LLM usage.

| comparison (control vs test) | sample size (n) (control/test) | time-in-workflow P-value (Δ; effect size) | LOS P-value (Δ; effect size) | turnaround-time P-value (Δ; effect size) |
|---|---|---|---|---|
| before vs after | 1165/884 | **0.003 (+77 s; 0.07)** | 0.878 (-1 day; 0.00) | **0.043 (-8.5 h; -0.08)** |
| before vs no LLM | 1165/676 | 0.240 (-42 s; -0.03) | **<0.001 (-1 day; -0.08)** | **0.032 (-10.5 h; -0.09)** |
| before vs LLM | 1165/209 | **0.000 (+498 s; 0.27)** | **<0.001 (+3 days; 0.19)** | 0.496 (-5.5 h; -0.03) |
| before vs BLEU20 | 1165/61 | **<0.001 (+436 s; 0.17)** | **<0.001 (+2 days; 0.12)** | 0.613 (-2.5 h; -0.02) |
| before vs BLEU50 | 1165/31 | **<0.001 (+694 s; 0.13)** | **<0.001 (+2 days; 0.09)** | 0.641 (+0.5 h; -0.02) |
| no LLM vs LLM | 676/209 | **0.000 (+421 s; 0.35)** | **0.000 (+4 days; 0.3)** | 0.370 (+5 h; 0.05) |

*Table 2: Statistical Analysis for Time-in-workflow. P-values are derived from Mann-Whitney U tests (α=0.05; two-tailed). Abbreviations: time-in-workflow = documentation time of a singular user per admission; LOS = length of stay of inpatient admission; turnaround-time = time between discharge and discharge letter sent; Δ = difference between medians of test groups; s = seconds; h = hours. Test groups: before = admissions before introduction of the LLM; after = admissions after introduction of the LLM; no LLM = admissions after introduction for which the LLM was not used; LLM = admissions for which the LLM was used; BLEU20 = admissions with a BLEU-4 score above 0.2; BLEU50 = admissions with a BLEU-4 score above 0.5. Cliff's Delta was used to measure effect size. A positive Δ indicates an increase compared to the control, and a negative Δ indicates a decrease compared to the control. Significant values are bold.*

Time-in-workflow of admissions for which the LLM has been used is increased, including admissions for which LLM-generated text has been copied, figure 6. When visualized on a user level, the complex nature of time-in-workflow is made clear. Figure 7 shows users spend varying amounts of time per note, which does not seem to be in relation to LOS.

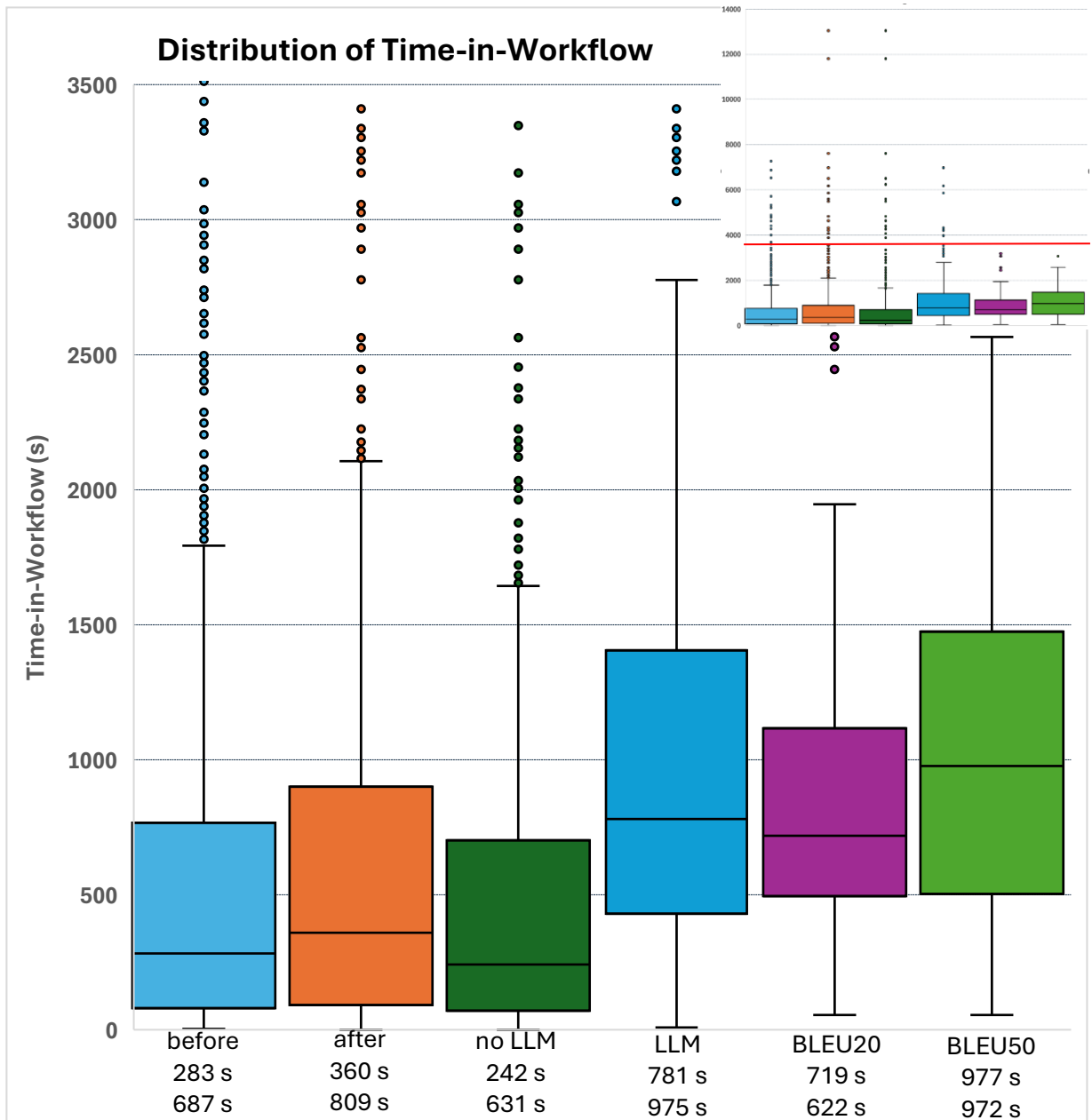

*Figure 6: Box Plots representing time-in-workflow per test group.* The y axis showing time-in-workflow in s = seconds is cut off at 3500 s for clarity. The x axis shows the test group names, medians, and interquartile ranges. s = seconds. Test groups: before = admissions before introduction of the LLM; after = admissions after introduction of the LLM; no LLM = admissions after introduction for which the LLM was not used; LLM = admissions for which the LLM was used; BLEU20 = admissions with a BLEU-4 score above 0.2; BLEU50 = admissions with a BLEU-4 score above 0.5. The box represents the middle 50% of the data, from lower quartile (bottom of box) to median (black line) to upper quartile (top of the box). The whiskers extend to the minimum and maximum of the data, points represent outliers. An inset shows the full distribution, the red line showing 3500 s.

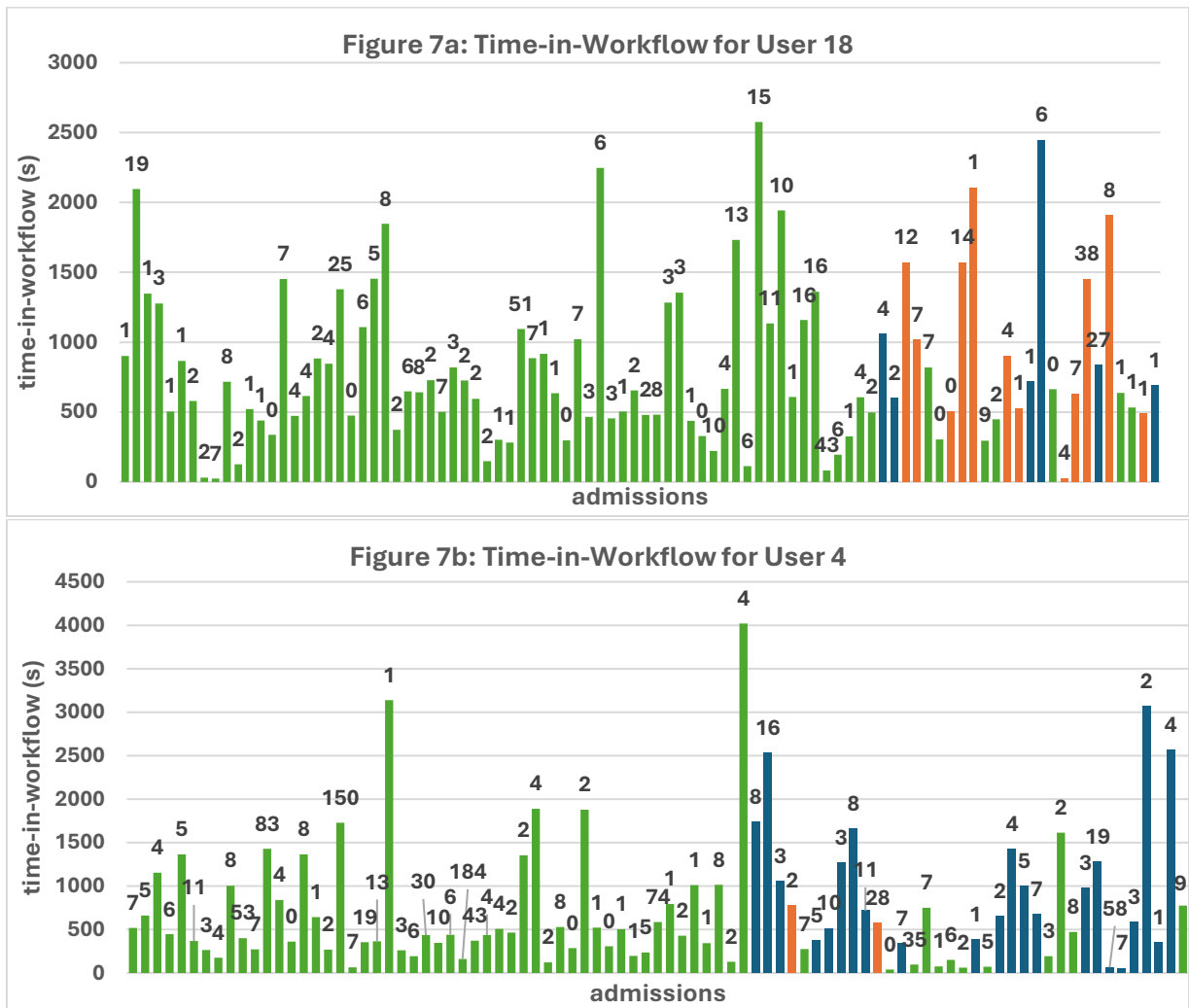

*Figure 7: Distribution of Time in workflow for two users.* Time-in-workflow (y axis) in s = seconds is shown for each admission (x axis) encountered chronologically (old to recent, left to right) between May 13$^{th}$, 2025, and September 14$^{th}$, 2025. The LOS is displayed for each admission on top of the column representing time-in-workflow. Admissions with an LLM-generated draft are coloured orange, and admissions with LLM-generated text in the final discharge letter (BLEU-4 >0,2) are coloured in blue.

## Discussion

This early clinical evaluation study sheds first light on the performance of LLM-generated discharge summaries in a Dutch tertiary care, academic hospital, providing quantifiable, reproducible measurements. User acceptance and feasibility, as measured by adoption and user experience, were high, paving the way for larger-scale implementation. High generation numbers and intent to use demonstrate HCPs in a tertiary academic medical centre perceived the need for LLM-assisted discharge documentation, and high copy counts and BLEU-scores suggest the LLM is adequate to this task. When the LLM-generated text is not fully retained in the discharge letter, this study demonstrates the LLM-generated draft can still influence the structure and content of the final discharge summary.

LLMs in simulated environments show clear potential to aid HCPs in clinical documentation tasks, generating medical documentation with similar quality to HCPs in less time [23,24]. This is underlined by the usage of real-world clinical data [24]. Generating discharge summaries, a time-consuming activity necessitating a lot of data extraction, has emerged as a promising early use-case for LLMs. Indeed, several studies prove that LLM-generated discharge summaries are largely non-inferior to HCP-written discharge summaries, while being more time-efficient [25–27,30,32]. Across medical research related to LLMs, translating potential to tangible clinical outcomes remains a challenge [33,34]. First publications describing the implementation of discharge summary generators are appearing, but quantifiable outcomes remain under-reported [29,37].

Adoption of the LLM varied between departments and seemed most useful to the surgery department, which demonstrated the highest BLEU-score averages combined with a high generation count and low generations-to-copy ratio. A low generations-to-copy ratio, as seen in the cardiology, gastroenterology and pulmonology departments, suggests the output of the LLM is well-adjusted to these departments. The internal medicine, paediatrics and neurology and neurosurgery departments needed more generations to achieve an LLM-generated discharge summary deemed adequate to be copied. The ENT and orthopaedics departments had similarly high generations-to-copy ratios, although their comparatively low generation counts suggest the need for department-specific prompting. Being the first point of in-hospital contact, the LLM did not encounter enough information in the patient files to generate meaningful summaries for the emergency department, which explains the low generation counts.

The BLEU-score was chosen to estimate the amount of LLM-text that was copied into the final letter. The copy/paste workflow encourages using large chunks of text from the LLM-generated draft, swapping single words or completely removing or adding new sentences. This, together with the inherent semantic similarity between LLM-generated and final discharge summary as they contain similar or identical medical information, can lead to over-estimation by metrics more indicative of semantic similarity such as the ROUGE or BERT scores [38].

The positive user experience and self-reported benefits back up the high adoption number and quantify the value of this EHR-integrated LLM. Choudhury et al. describe that ease of use and perceived usefulness of an LLM have a positive direct effect on trust in an

LLM, which in turn has a positive direct effect on intent to use an LLM [39]. Our study reflects this in the similar perceived usefulness (self-reported time benefit) and intend to use percentages. However, the importance of perceived usefulness and ease of use seems to exceed trust in influencing intent to use in our study. This raises the idea of a 'healthy distrust', or 'healthy suspicion', in the attitudes of HCPs towards this LLM. This attitude has already been described as an 'ambivalent' attitude towards GenAI at large [40]. This could explain why users tended to distrust the LLM, but not their colleagues. The high adoption numbers achieved despite low trust among users further strengthens the study.

In an effort to provide a valuable extrinsic validation, we measured the documentation time for each admission encountered by the users. However, as shown by figure 7, this data is influenced by the complex and cooperative nature of inpatient care in an academic hospital. Multiple residents or HCPs will be editing the discharge note or letter of a given admission, before being reviewed by a specialist. Additionally, junior doctors or medical students may contribute to this workflow. This results in a lot of 'noise' attributed to admissions for which the users opened the discharge documentation without making significant contributions, for instance to review, to fill in for a colleague, or to contribute to a common discharge note.

Additionally, time was measured from first edit action to saving of the document, not accounting for potential disruptions in workflow. The User Action Logging (UAL) data was not granular enough to warrant exclusions, thereby compromising the analysis we performed. It is also important to mention that this analysis was performed purely with exploratory intent, as the user instructions warranting safety effectively counter-acted any reductions in documentation time.

Thus, it was expected that the LLM did not yet decrease documentation time. This is a finding reflected by previous studies describing a lack of time benefits in the initial phases of LLM implementations [38,41,42]. However, demonstrating extrinsic benefits is necessary for widespread acceptance of GenAI features and currently missing in the literature [34]. Implementing and describing reliable, accurate and scalable metrics for extrinsic validation remains a challenge to be addressed in this field of research.

The lack of extrinsic time benefits demonstrated by this analysis lies in contrast to the high subjective reports of time benefits. As the primary objective of EHR-integrated LLMs should be to reduce administrative burden for HCPs, demonstrating subjective time benefits, or perceived usefulness, is a promising result considering the workflow of this EHR-integrated LLM is not optimised yet.

We chose to report several outcomes in this study to underline the complexity of implementing and evaluating an LLM in clinical practice, thereby limiting the depth in which these outcomes can be discussed. Many important aspects, such as the evaluation and management of erroneous LLM outputs, lie outside the scope of this pilot study.

The outcomes measured and reported in this study are additionally prone to confounding factors. As discussed above, the time-in-workflow measurements represent only a fraction of a complex process and do not account for possible interruptions. Additionally, time-in-workflow can be confounded by many other factors, such as documentation and EHR-use related skills or cognitive load. Similarly, adoption and user experience can be influenced by AI literacy and willingness to adopt new technologies [43,44].

This early clinical evaluation of an EHR-integrated LLM generating discharge summaries in a Dutch academic tertiary care hospital demonstrates high user acceptance and feasibility, paving the way for phase 2 implementation of this LLM-use-case. Based on the outcomes reported here, our medical centre approved further implementation of this EHR-integrated LLM. This will enable future research focused on extrinsic validation which will have to address the challenges related to accurate time measurements.

## Methodology

**Setting and participants**

This was a prospective mixed methods pilot study on the phase 1 implementation of an EHR-integrated LLM-driven functionality generating discharge summary drafts, conducted according to the TRIPOD-LLM reporting standards [45]. We defined phase 1 implementation to identify barriers to adoption and improve user experience with a small, selected user group. This use-case has previously passed the phase 0 non-production validation, ensuring sufficient quality to warrant usage in production. The next step, phase 2 implementation, will be performed on a larger cohort and focus on extrinsic evaluation.

From July 14th to September 14th, 2025, 25 pilot users (21 residents and 4 physician assistants) across 10 departments in our Dutch tertiary academic hospital tested this LLM application in daily clinical practice in 253 admissions, achieving a total of 379 generated discharge summaries. Recruitment was done via email and presentations during daily briefings until sufficient users were reached. All users were instructed personally and had to complete a mandatory custom-made e-learning available in the supplementary information.

**Outcomes**

Primary outcomes were adoption and user experience. Potential time benefits related to LLM usage were explored as a secondary outcome.

Adoption was measured on two different levels. Adoption of the LLM was based on 'clicks', such as requesting and copying an LLM-generated draft, and the number of successful generations achieved. Text adoption of the LLM was calculated between the final discharge letter text and the LLM-generated draft text. BLEU scores were used to calculate the percentage of identical words (or n-grams) between the discharge summary from the final discharge letter and the LLM-written discharge summary. 10 cases were randomly selected to estimate how closely the BLEU-scores approximated the amount of text copied from the LLM-generated draft. For these 10 cases, the percentage of words in the discharge summary from the final discharge letter originating from the LLM-draft was manually calculated.

User experience was expressed in pre-determined concepts: perceived usefulness, trust, ease of use, and intent to use[39,46]. A link to the survey in Dutch can be found in the supplementary information. The survey was filled in after completion of the pilot study.

Potential time benefits were investigated using User Action Logging (UAL) data from the EHR. The time spent working in discharge notes and letters was added up per unique admission (time-in-workflow) for all users and subsequently divided into 'before' (13th of May to 13th of July 2025) and 'after' (14th of July to 14th of September 2025) access to the LLM. The admissions 'after' were further divided as shown in figure 8. Time-in-workflow was measured from the first action in a document to the moment the document is saved, in seconds. Idle intervals could not be excluded, nor were any thresholds applied to the data. LOS and turnaround-time were added to each admission.

**BLEU-score Calculations**

The sentence_ bleu package from the Natural Language Toolkit (NLTK) library was used to estimate the amount of LLM-generated text present in the final discharge letter. The BLEU-score measures the overlap of n-grams between a reference (the LLM-generated text) and a candidate (the final discharge letter) text, expressed in a score ranging from 0 (no identical n-grams) to 1 (all n-grams are identical), with an n-gram representing a sequence of words. For instance, a BLEU-score with an n-gram of 4 (BLEU-4 score) measures the number of times four subsequent words are identical in a candidate text compared to a reference text. We used n-grams of 1, 2 and 4. A brevity penalty is applied when the candidate text is shorter than the reference, and Laplace smoothing (method1) was performed. The underlying code for this analysis is shown in supplementary figure 7.

Citations were removed from the LLM-generated drafts prior to the calculations. The final discharge letter text was pre-processed using a VBA script in Microsoft Excel (Microsoft, Seattle, WA). This script removed most text not belonging to the discharge summary or conclusion of the letter. All cases were manually controlled.

**Time Measurements**

UAL data from the EHR was retrieved for the extrinsic validation for the study period and the two months prior (May 13[th] to September 14[th], 2025), filtering for discharge notes and discharge letters. All admissions for which the users edited a discharge note or letter in the given timeframe were included for the analysis, as shown in figure 8. This data includes time-in-workflow, LOS and turnaround-time. Not all admissions with an LLM generation could be retrieved from the UAL database. From the 253 admissions with an LLM generation, 17 (7%) were missed by the final UAL search. For the extrinsic validation, another 6 (2%) were excluded for missing time measurement, 4 (1.5%) were excluded for not being discharged by the end of the study period, and 17 (7%) were excluded for not returning time measurements for a user, resulting in the 209 admissions shown in figure 8.

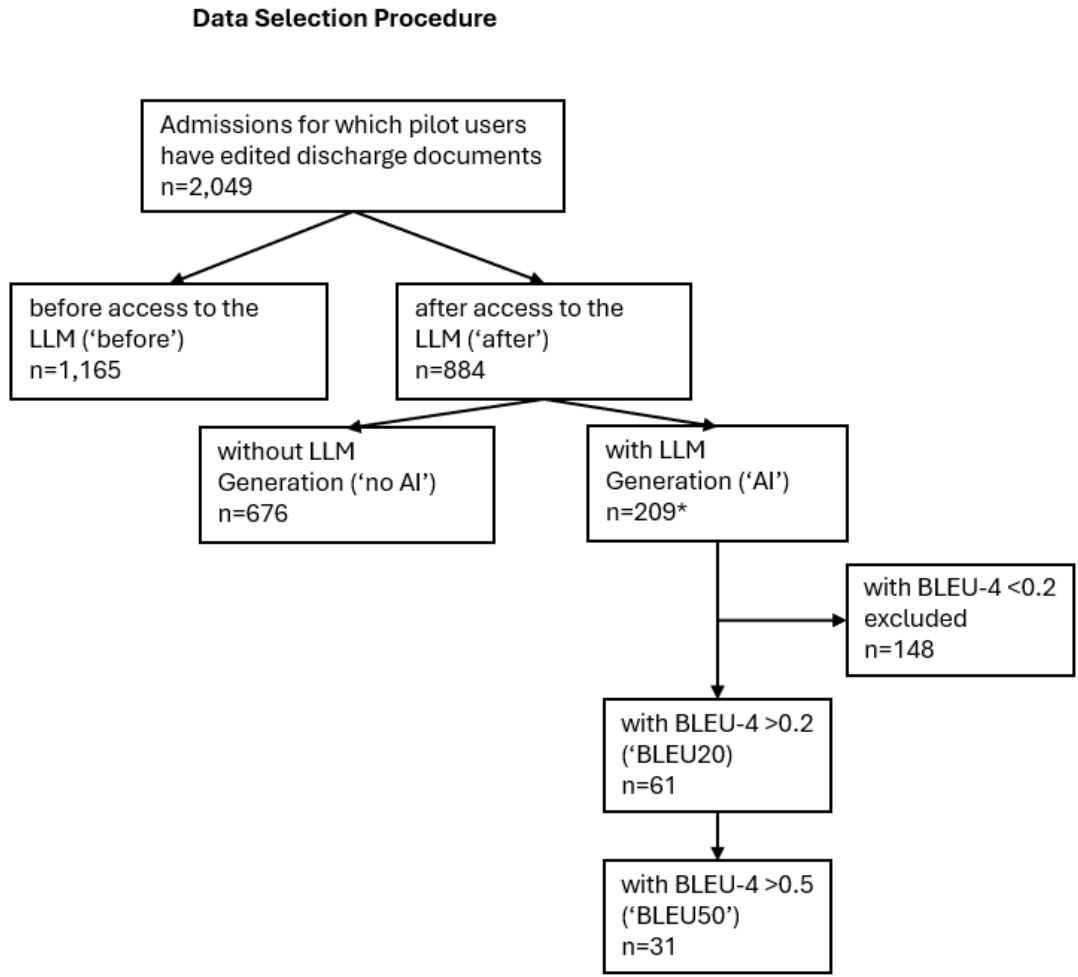

*Figure 8: Data selection procedure for the extrinsic validation. Each box contains a dataset. Each arrow represents extraction of a subset of data from a database or dataset. \*One admission with an LLM-generated draft was discharged on July 12<sup>th</sup>, but the discharge letter was only sent on September 17<sup>th</sup>, 67 days later.*

**Statistical Analysis**
The statistical analysis was performed on SPSS Statistics 28 (IBM, Armonk, NY). The Kolmogorov-Smirnov and Shapiro-Wilk tests of normality and Spearman's rho (two-tailed) were used to investigate the distribution and correlation of time-in-workflow, LOS and turnaround-time. The 'before' group was compared to the 'after' group, and all subgroups hereof, through a Mann-Whitney U test (α=0.05; two-tailed).

**LLM Specifications and Data**
The study was started on the GPT-4o LLM and switched to the GPT 4.1 LLM on August 6<sup>th</sup>, 2025 (OpenAI, San Francisco, CA). The LLMs were run on the Microsoft Azure Cloud Service (Microsoft, Seattle, WA) and integrated into the EHR (Epic Systems, Verona, WI). Summaries were generated with a maximum context of 128,000 and a maximum output of 4,096 tokens (GPT-4o) and 32,000 tokens (GPT-4.1). The prompt to the model used clinical information extracted from the electronic health record in Dutch. This included demographics information, such as name, age, gender, and problem list information, and notes information including any linked attestations.

A standardized prompt directed the LLM to write a summary of the hospital admission and used a pre-loaded demographics string as the first sentence. The first prompt section outlined the format, target audience, and other characteristics of the requested summary depending on the requested summary type. It directed the LLM to focus on the key reason for admission and its pertinent co-morbidities. The next section gave instructions on data formats and medical terminology as well as instructions which data to prioritize or avoid. The prompt closed with instructions to be concise, not to fabricate information, and not to draw conclusions not present in the source notes. The prompt was largely identical for all cases, and changes performed were minor, for instance a shift from 'diagnose' to 'problem'.

Any pre-existing discharge summary was intentionally withheld to evaluate the model's ability to summarize on its own. The model was instructed to output the summary in Dutch. We used a temperature of 0 and top P of 1 for decoding parameters. No fine-tuning was performed. Summaries were generated without post-processing.

The datasets analysed during the current study contain sensitive patient information and are therefore not publicly available. Data can be made available upon reasonable request to the corresponding author.

**Clinician-AI Interaction**

A button leading to the LLM was added to the toolbar in the EHR for the pilot users. Once an admission was chosen, the interface opened in the sidebar of the chosen patient file in the EHR. Output format (narrative, problem-oriented, system-oriented) and length (unspecified, <10 sentences, 10-20 sentences, 20-30 sentences) settings were chosen here. The default setting was narrative summary of unspecified length. The LLM generated a discharge summary on demand. Feedback could be left in the EHR after a generation was completed. References to the source text were given for each piece of information. If copied, the LLM-generated text was moved to the clipboard without the references and could subsequently be pasted into and edited in the discharge note or letter. Screenshots of the user interface can be found in the e-learning provided in the supplementary information.

The users were instructed to click the 'generate' button and switch to their usual workflow while the LLM was 'thinking'. Users were instructed to make a discharge summary draft themselves before reading the LLM-generated draft to mitigate potential omissions from the LLM.

**Ethical Considerations**

This prospective study has been approved by our institutional review board according to the declaration of Helsinki (ref. M24.328217). All data was encrypted in transit to and from the LLM, ensuring compliancy with Dutch privacy laws (WPG). We determined this LLM-use case not to be an AI-based clinical decision support system (AICDSS), and therefore not to fall under the Medical Device Regulation.

All users participating in this study gave informed consent and acknowledged their responsibilities related to testing a functionality in development. All LLM-generated content was reviewed and approved by an HCP and was only used in the discharge summary when deemed adequate. No obvious bias or harm was detected in the LLM-generated output.

## Acknowledgements

We gratefully acknowledge the time, effort and enthusiasm of the participating residents and physician assistants willing to test the LLM. We thank Gerhard Bultema, application specialist at our hospital, for helping with technical issues related to the implementation of the LLM. Ben Weinstein, Epic Systems, played a crucial role in the data extraction necessary for this study.

## Author Contributions

N.N. performed the recruitment and support of the pilot users and the data analysis. The study was designed with the help of T.M., C.B. and R.S. N.N. performed the redaction of the manuscript under supervision of C.B., J.D. and R.S. Revision of the manuscript before submission was done by T.vdL. and J.O.

## Competing Interests

The authors have no competing interests to declare. This research did not receive funding.